\documentclass[runningheads]{llncs}
\usepackage[T1]{fontenc}
\usepackage{graphicx}
\usepackage[english]{babel} 

\usepackage{listings}
\usepackage{ragged2e}
\usepackage{float}
\usepackage{makecell} 
\usepackage{fontawesome5}
\usepackage{upquote}

\begin{document}
\title{Can LLMs Assist Computer Education? an Empirical Case Study of DeepSeek}
\titlerunning{LLMs Assist Computer Education}
%
%
\author{Dongfu Xiao\inst{1}$\star$ \and Chen Gao\inst{1}$\star$ \and Zhengquan Luo \inst{1} \and Chi Liu \inst{1}\faIcon{envelope} \and Sheng Shen \inst{2}}
\authorrunning{D. Xiao et al.}
%
\institute{Faculty of Data Science, City University of Macau, Macao SAR, China \and
Design and Creative Technology Vertical, Torrens University Australia, NSW, Australia \\
$\star$ Equal contribution \\
\faIcon{envelope} Corresponding author: \email{chiliu@cityu.edu.mo}}


%
\maketitle              
\begin{abstract}
This study presents an empirical case study to assess the efficacy and reliability of DeepSeek-V3, an emerging large language model, within the context of computer education. The evaluation employs both CCNA simulation questions and real-world inquiries concerning computer network security posed by Chinese network engineers. To ensure a thorough evaluation, diverse dimensions are considered, encompassing role dependency, cross-linguistic proficiency, and answer reproducibility, accompanied by statistical analysis. The findings demonstrate that the model performs consistently, regardless of whether prompts include a role definition or not. In addition, its adaptability across languages is confirmed by maintaining stable accuracy in both original and translated datasets. A distinct contrast emerges between its performance on lower-order factual recall tasks and higher-order reasoning exercises, which underscores its strengths in retrieving information and its limitations in complex analytical tasks. Although DeepSeek-V3 offers considerable practical value for network security education, challenges remain in its capability to process multimodal data and address highly intricate topics. These results provide valuable insights for future refinement of large language models in specialized professional environments.
\end{abstract}

\section{Introduction}
Recent advancements in large language models (LLMs) have opened substantial opportunities to transform professional education in computer networking and security. These models exhibit advanced capabilities in processing and articulating intricate technical concepts\cite{xu2024large,zhang2025llms}, making them potent educational tools that can complement conventional learning methods\cite{raihan2025large}. Their capacity to produce precise, context-aware explanations of networking and security principles\cite{hassanin2024comprehensive}, coupled with the generation of tailored instructional content\cite{liu2024large}, establishes LLMs as valuable resources for developing professional expertise. Although initial studies have shown LLMs' proficiency in applying domain-specific knowledge, their efficacy in comprehensive networking education—particularly concerning complex technical concepts and practical problem-solving scenarios—necessitates systematic evaluation using established professional benchmarks\cite{xu2024large}.

Notably, next-generation LLMs like DeepSeek-V3 present significant untapped potential in this domain. Built on the Transformer architecture, DeepSeek-V3 has achieved groundbreaking progress in natural language processing through large-scale data training, demonstrating unique advantages in code generation and mathematical reasoning. Its 128K context window and dynamic knowledge retrieval capabilities are particularly promising for handling complex networking tasks with long-sequence dependencies, including protocol analysis and topology configuration \cite{liu2024deepseek}. This combination of technical features suggests new possibilities for advancing both research and education in computer networking.

To rigorously evaluate the practical effectiveness of DeepSeek-V3 in computer networking, a systematic assessment across diverse network scenarios is necessary. For this study, we selected two authoritative test datasets: the latest simulated question bank from the Cisco Certified Network Associate (CCNA) exam—a widely recognized industry standard—and a subset of real questions from China's 2022-2023 Network Engineer certification (part of the Computer Technology and Software Professional Qualification Examination, commonly known as the Soft Exam) \cite{cisco2023,ruankao2023}.

The CCNA exam covers fundamental networking concepts, including IP addressing, routing, and switching, while the Network Engineer exam assesses broader competencies such as network planning, security, and management. Both examinations integrate theoretical networking principles with practical problem-solving, presenting challenging questions that test both conceptual and applied knowledge. 

Given their comprehensive coverage, technical depth, and varying difficulty levels, these exams serve as robust benchmarks for evaluating DeepSeek-V3’s capabilities in computer networking. By analyzing the model’s performance on these standardized assessments, we can assess its proficiency in understanding, applying, and solving network-related problems. The findings will provide valuable insights for further optimization and real-world deployment of this language model in networking applications.

\section{Related Works}
\subsection{LLM}
LLMs constitute a category of deep neural networks characterized by their massive scale (typically comprising billions to trillions of parameters), which demonstrate exceptional proficiency in language comprehension and generation, thereby enabling seamless human-machine interaction through natural language interfaces \cite{bommasani2023holistic}. The pervasive adoption of LLMs in recent years has driven significant progress across multiple disciplines, including healthcare \cite{ting2024performance,alfertshofer2024sailing}, educational technology \cite{alqahtani2023emergent}, computer science research \cite{hou2024large}, and cybersecurity \cite{donadel2024can}. However, these performance improvements have been accompanied by exponentially increasing computational costs during training. Current state-of-the-art models exhibit substantial variations in training expenditures: Google's Gemini Ultra ranks as the most expensive at \$191 million, followed by OpenAI's GPT-4 at \$78 million. Notably, DeepSeek achieves competitive performance with significantly lower training costs of only \$5.6 million \cite{mikhail2025performance}. Among existing LLMs, DeepSeek stands out for its exceptional cost-efficiency while remaining open-source, contributing to its widespread adoption across various application domains.

\subsection{Evaluation of LLMs in Professional Domains}
The application of LLMs to standardized and professional certification exams has grown significantly across multiple disciplines. This systematic assessment examines their competence in domain-specific knowledge evaluation, reasoning precision, and practical utility \cite{alfertshofer2024sailing,mendoncca2024evaluating,liu2024performance}. Current research in this area reveals a strong emphasis on medical applications, particularly in evaluations such as Medical Licensing Examinations \cite{alfertshofer2024sailing} and Nuclear Medicine Physician Board Examinations \cite{ting2024performance}. In computer science, \cite{mendoncca2024evaluating} analyzed ChatGPT-4's performance on Brazil's National Undergraduate Computer Science Examination.

The predominant evaluation approach involves curating authoritative exam questions from standardized tests and benchmarking LLMs' ability to answer them. Assessments commonly employ single-choice and multiple-choice formats to rigorously measure domain-specific problem-solving capabilities, quantified by accuracy in addressing discipline-related queries \cite{mendoncca2024evaluating}. Additionally, multiple studies have explored LLM performance across different languages \cite{alfertshofer2024sailing,ting2024performance,liu2024performance}. Nevertheless, no existing work has conducted a systematic evaluation of DeepSeek's performance on network engineer certification examinations.

\section{Methodology}
\subsection{DeepSeek}
DeepSeek's language model achieves state-of-the-art performance in natural language processing, with empirical results demonstrating robust capabilities in both text generation and complex reasoning tasks.

Benchmark evaluations confirm DeepSeek-V3's advanced language and reasoning capabilities, with officially reported accuracies of 92.3\% on GSM8K mathematical reasoning (5-shot), 79.6\% on English AGIEval (0-shot), and 90.1\% on Chinese C-Eval (5-shot) \cite{deepseekReport}. These results, derived from standardized evaluations under controlled protocols, position the model among the top-performing open-weight systems and validate its reliability for this study's methodology.

This study employs the official DeepSeek API for experimental analysis \cite{deepseek2024}. The DeepSeek-V3 model utilizes a mixture of experts (MoE) architecture comprising 671 billion total parameters with approximately 37 billion activated per token \cite{liu2024deepseek}. Pretrained on 14.8 trillion tokens of curated multilingual data from web texts, technical documents, and academic publications, it demonstrates robust performance across diverse domains while maintaining efficient inference \cite{liu2024deepseek}.

\subsection{Question Selection and Classification}
To systematically assess large language models' capabilities in computer network security, we selected two authoritative test banks: the official question repository from China's Software Professional and Technical Qualification Examination (200 questions total) and the CCNA simulation test bank (380 questions total). Since DeepSeek-V3 primarily processes textual data, we excluded image-based questions, resulting in a final evaluation set comprising 122 questions from the Chinese examination and 199 from the CCNA test bank. 

These curated question sets strictly align with the respective examination syllabi, ensuring comprehensive coverage of fundamental computer network security concepts. The selection process guarantees that the evaluation encompasses the core knowledge domains essential for network security proficiency.

To evaluate the relative difficulty of both examinations, we employed system prompts to elicit DeepSeek-V3's difficulty ratings for each question, subsequently computing the mean difficulty score for each test.

\begin{figure}
\centering 
\includegraphics[width=\textwidth]{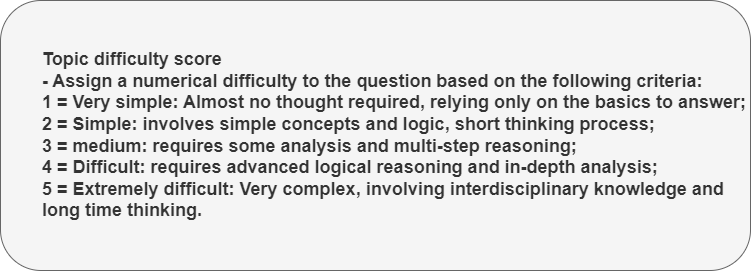} \caption{Difficulty Assessment Prompt Engineering} 
\label{fig:difficulty_assessment} 
\end{figure}

The analysis revealed comparable difficulty levels between the examinations for single-choice questions, with average difficulty scores of 2.1 and 2.2 respectively, indicating statistically equivalent challenge levels.(see Figure \ref{fig:difficulty_assessment})

Based on the question-selection methods, the questions are classified into single choice and multiple choice types \cite{bommasani2023holistic,liu2024performance,mendoncca2024evaluating}. The single-choice questions consist of three distractors and one correct answer, whereas the multiple-choice questions include three distractors with two correct responses. For more granular analysis, DeepSeek-V3 categorized both question types into six thematic domains based on the official CCNA classification framework.

\subsection{Prompt Engineering}
Given the substantial influence of prompt engineering on the output of LLMs, rigorous standardization of the model output is implemented \cite{donadel2024can}. The question-selection methodology requires specifying the exact number of correct answers for each item. Based on the question language, the model is directed to provide explanations in either Chinese or English. Questions and corresponding answers are submitted through the model's API interface. The experimental framework employs the following role definitions(Figure \ref{fig:prompt_engineering}):  

\begin{figure} 
\centering 
\includegraphics[width=\textwidth]{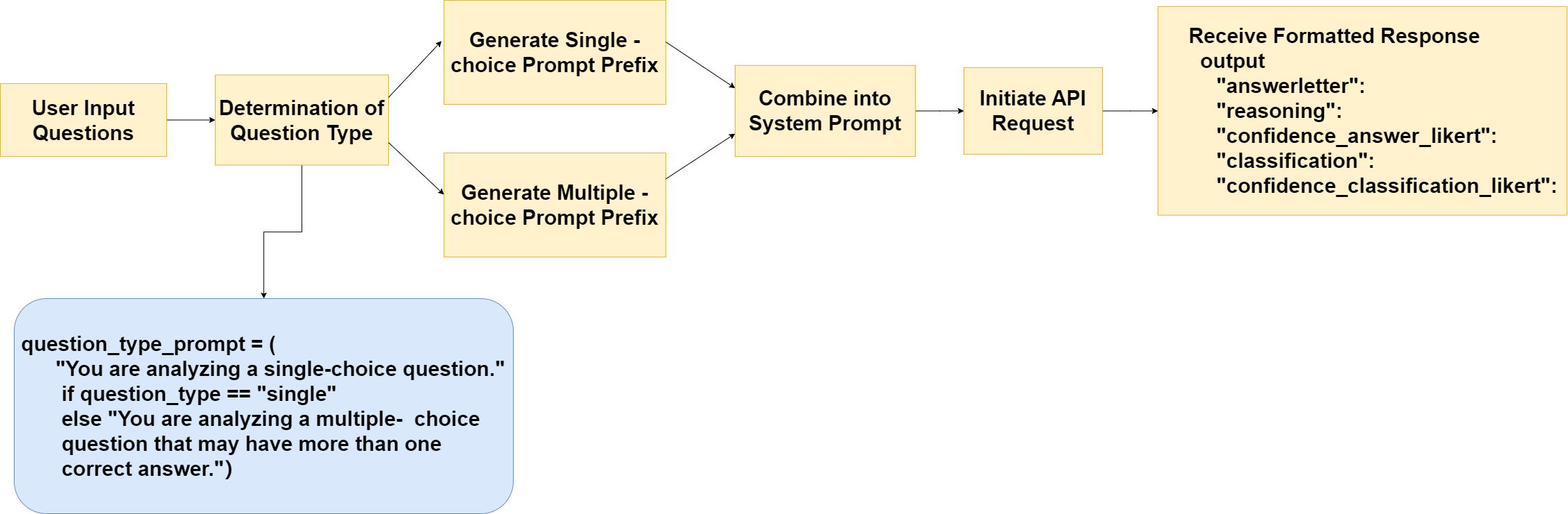} \caption{Prompt Engineering} 
\label{fig:prompt_engineering} 
\end{figure}

\begin{lstlisting}[
    basicstyle=\ttfamily\footnotesize, 
    breaklines=true, 
    columns=flexible  % 让字符间距更紧凑
]
system_prompt = f'''
{question_type_prompt}
You are an experienced professional in the field of computer networks, 
possessing deep theoretical and practical knowledge. 
You are taking a computer network examination composed of {question_type}-choice questions, 
requiring in-depth professional knowledge and analytical skills.
Please answer each question according to the following requirements, 
and ensure all outputs are in JSON format.
Your output must be as follows (ensure the field names match exactly):
{{
  "answerletter": "Your answer",
  "reasoning": "Your explanation",
  "question classification": "Your classification"
}}
Question Classification:
- Classify the question into one of the following two categories and return the corresponding string:
- "Lower-order": Questions that test memory and basic understanding.
- "Higher-order": Questions that require applying knowledge, analyzing abilities, or evaluation.
'''
\end{lstlisting}

\subsection{Data Collection and Assessment}
The responses generated by DeepSeek-V3, including its answers, detailed explanations, and question classifications, were systematically collected and processed. For performance evaluation across both the CCNA certification and China's National Computer Technology and Software Professional Qualification Examination (Network Engineer), only single-choice questions were considered. This selection criterion was uniformly applied, as all questions in the Network Engineer examination question bank were exclusively single-choice format, while for CCNA, single-choice questions were specifically selected from the available pool.

\subsection{Multi-dimension Evaluation}
\subsubsection{Role Dependency}
To investigate the actual impact of prompt engineering on large language model outputs, this study introduces a modified prompt version that eliminates role specifications. Rather than predefining the model as a "computer network expert," this approach relies solely on the provided information and task requirements to guide the model's responses.  

\subsubsection{Cross-linguistic Reliability}
To evaluate DeepSeek-V3's cross-lingual processing capabilities, this study employed question banks from both the CCNA certification and China's National Computer Technology and Software Professional Qualification Examination (Network Engineer level). The experimental methodology involved creating parallel Chinese-English versions of these examination questions through professional translation (see: Chinese-English translation methods). Comparative analysis of model performance on translated versus original-language questions provided quantitative assessment of its cross-lingual adaptation capabilities.

\subsubsection{Reproducibility of Answers}
A total of 50 questions are randomly selected at a time from three question banks: the CCNA single-choice question bank, the CCNA multiple-choice question bank, and the soft exam single-choice question bank.For each of the selected questions, 50 independent questioning operations are carried out \cite{reproducibility}. Record the frequency of each answer for each question. Among all inquiries, if the proportion of the same answer reaches 75\% or more, it is regarded as an answer with high reproducibility[Literature: Reproduction]. After the analysis, compare the answers with high reproducibility with those with low reproducibility.

\subsection{Statistical Analysis }
To evaluate performance, the chi-square ($\chi^2$) test was applied when expected frequencies exceeded 5, and Fisher's exact test was used for smaller frequencies.  Analyses included Role-Independent Performance, cross-language performance, Performance by Question Type, and Performance by Topic.  Lastly, Reproducibility of Responses was assessed using the same methodology.  For CCNA and Network Engineer exam questions, repeated experiments (five iterations per set) were conducted to derive p-values, thereby reinforcing the validity, reliability, and reproducibility of the findings across all evaluation dimensions.

Data analysis was conducted using Python (version 3.13.1) with libraries such as Pandas (version 2.2.3), NumPy (version 2.2.1), and SciPy (version 1.15.0).  These tools facilitated comprehensive statistical evaluations and ensured accurate computation.

\section{Results }

\subsection{Overall Performance}
In the CCNA exam domain, the model demonstrated an accuracy of 87.4\% (145/166), whereas its performance in the Network Engineer exam domain yielded an accuracy of 82.0\% (100/122). A chi-square test comparing performance in these two domains resulted in a p-value of 0.0386 ($p < .05$), indicating that the difference observed is statistically significant. The analysis presented above pertains specifically to single-select questions. To further assess the model’s proficiency, a separate evaluation was conducted on multi-select questions within the CCNA exam domain. In this subset, the model attained an accuracy of 81.8\%, successfully answering 27 out of 33 multi-select items. (refer to Table \ref{tab:performance_summary})

\begin{table}[h!]
    \centering
    \caption{Overall of DeepSeek-V3 Performance Across Exam Domains and Formats}
    \label{tab:performance_summary}
    \renewcommand{\arraystretch}{1.5}
    \resizebox{\textwidth}{!}{
            \setlength{\tabcolsep}{8mm}{
    \begin{tabular}{l|c|c}
    \hline
    \textbf{Exam Domain}        & \textbf{Format}               & \textbf{Accuracy (\%)}  \\ \hline \hline
    CCNA                        & Single-select                 & 87.4 \textnormal{(145/166)}                        \\ \hline
    Network Engineer            & Single-select                 & 82.0 \textnormal{(100/122)}                        \\ \hline
    CCNA                        & Multi-select                  & 81.8 \textnormal{(27/33)}                          \\ \hline
    \end{tabular}}}
\end{table}

\begin{figure}
\centering 
 \includegraphics[width=0.95\textwidth]{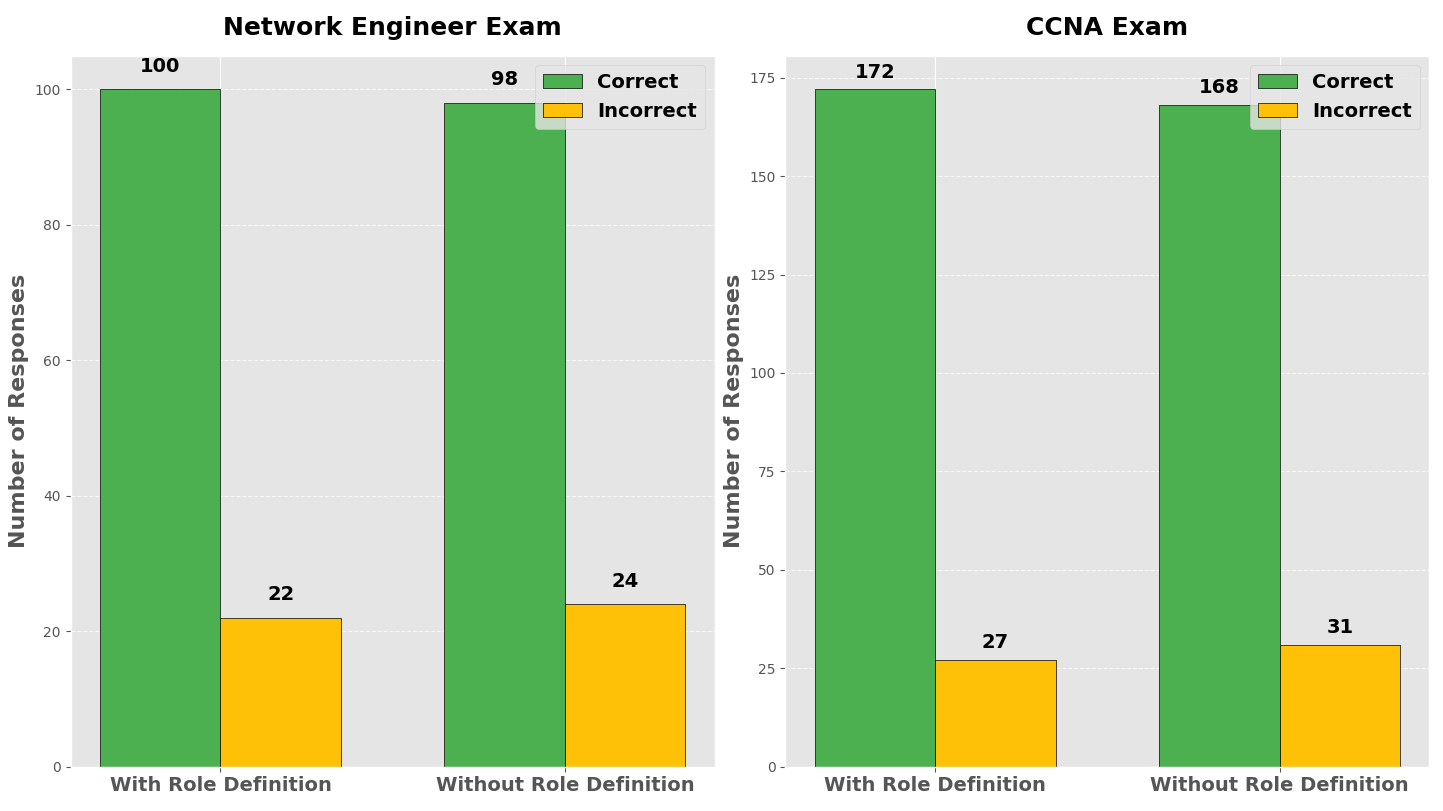} \caption{Impact of Prompt Design on Exam Answers} 
\label{fig:overll} 
\end{figure}

\subsection{Role Dependency Performance} In the Network Engineer exam domain, the model achieved an accuracy of 81.9\% (100/122) under the "With Role" condition, compared to 80.3\% (98/122) in the "Without Role" condition ($p = .88$), indicating no statistically significant difference (Fig \ref{fig:overll}). Similarly, within the CCNA exam domain, the model demonstrated an accuracy of 86.4\% (172/199) with role specification, compared to 84.4\% (168/199) in the "Without Role" baseline condition ($p = .53$) (Fig \ref{fig:overll}), again showing no notable variation. These findings suggest that incorporating role definitions within the prompt does not substantially influence the performance of DeepSeek-V3 in certification evaluations.

\subsection{Cross-language Performance} DeepSeek-V3 demonstrated consistent accuracy across both the original and translated versions. In the Network Engineer exam domain, the model achieved an accuracy of 82.0\% (100 out of 122) for the original version, compared to 83.6\% (102 out of 122) for its translated counterpart ($p = .16$), as shown in Table \ref{tab:cross_lingual}. Similarly, for the CCNA domain, the model attained an accuracy of 86.4\% (172 out of 199) for the original version and 83.0\% (165 out of 199) for the translated version ($p = .20$), as detailed in Table 2. These findings indicate that the translation process does not significantly affect model performance.

\begin{table}[ht]
    \centering
    \caption{Cross-language Performance Comparison Between Original and Translated Question Sets}
    \label{tab:cross_lingual}
    \renewcommand{\arraystretch}{1.5}
    \resizebox{\textwidth}{!}{
            \setlength{\tabcolsep}{1mm}{
    \begin{tabular}{l|l|l|l}
    \hline
    \textbf{Exam} & \makecell{\textbf{Original} \\ \textbf{Accuracy (\%)}} & \makecell{\textbf{Translated} \\ \textbf{Accuracy (\%)}} & \makecell{\textbf{Statistical Significance} \\ \textbf{($\chi^2$)}} \\
    \hline
    \hline
    Network Engineer & 82.0 (100/122) & 83.6 (102/122) & \makecell{$p = .16$} \\  
    \hline
    CCNA & 86.4 (172/199) & 83.0 (165/199) & \makecell{$p = .20$} \\  
    \hline
    \end{tabular}}}
\end{table}

\subsection{Performance by Question Type} 
\begin{figure}[H]
\centering 
\includegraphics[width=0.7\textwidth]{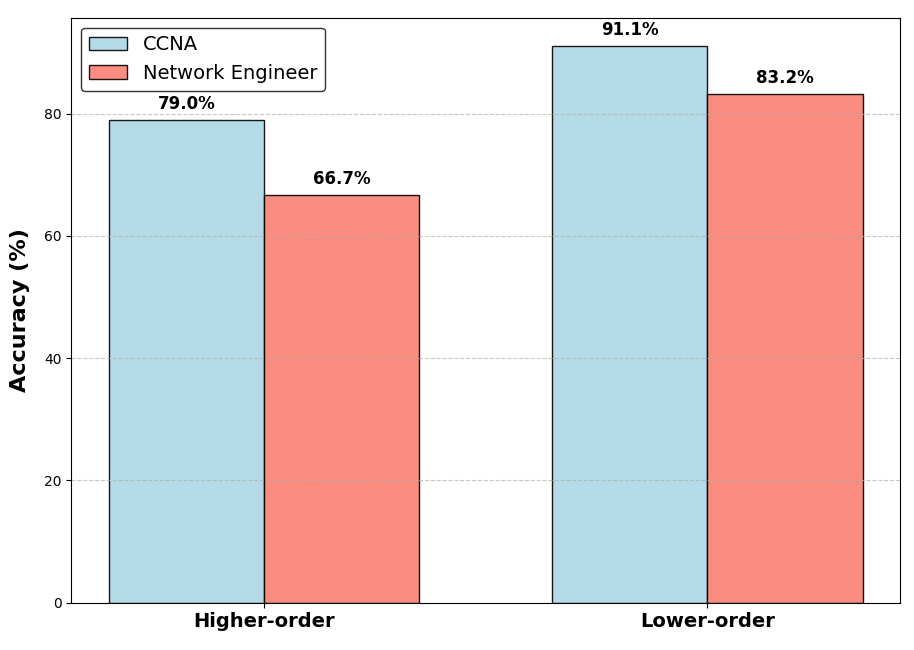} \caption{Accuracy Comparison: Higher-order vs Lower-order} 
\label{fig:accuracy_gap} 
\end{figure}

The DeepSeek-V3 model exhibited a pronounced accuracy gap between higher-order and lower-order questions. For the CCNA exam assessment, the model demonstrated a 12.1\% accuracy gap: 91.1\% on 123 lower-order questions, compared to 79.0\% on 76 higher-order items. This disparity was statistically significant ($\chi^2$ test, $p < .001$). In the Network Engineer exam evaluation, the model exhibited a 16.5\% accuracy gap: achieving 83.2\% accuracy on 107 lower-order questions, while higher-order items (15 total) reached only 66.7\%. This performance divergence was also statistically significant ($\chi^2$ test, $p < .05$). These findings underscore the model's pronounced advantage in factual recall tasks relative to complex reasoning scenarios (Fig.\ref{fig:accuracy_gap}).

\subsection{Performance by Topic}
\begin{figure}
\centering 
 \includegraphics[width=0.9\textwidth]{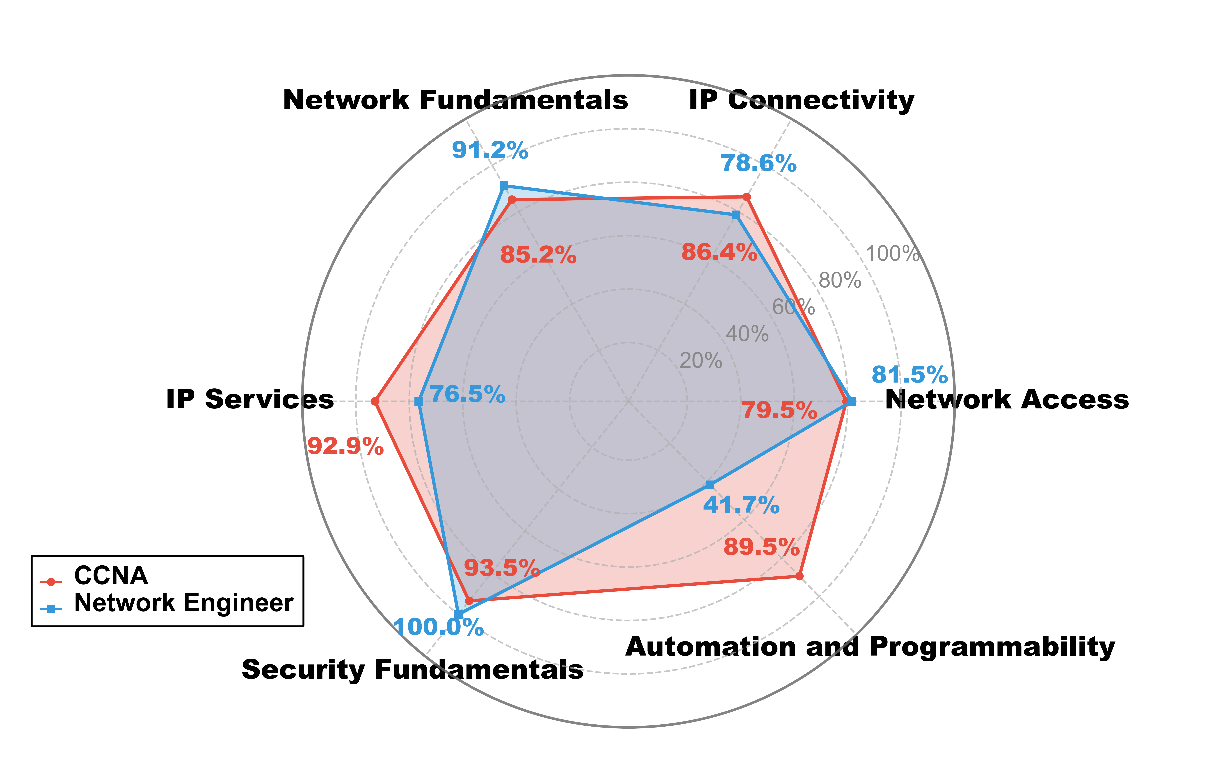} \caption{Deepseek-V3's Performance across Diverse Topics} 
\label{fig:topic_performance} 
\end{figure}

Figure \ref{fig:topic_performance} illustrated that DeepSeek-V3 achieved high precision in the field of computer network security across two test papers. A comprehensive analysis revealed that DeepSeek-V3 performed better on the CCNA questions than on those from the Network Engineer Exam. A substantial discrepancy in accuracy was observed between the two sets of questions for the themes of Automation and Programmability and IP Services, with the model attaining accuracy rates of 92.9\% and 89.5\% on the CCNA test questions, compared to 76.5\% and 41.7\% on the Network Engineer test questions. Statistical analysis indicated no significant difference in DeepSeek-V3’s theme-based performance in the CCNA exam, as robust results were achieved across all themes; however, in the Network Engineer exam, DeepSeek-V3’s performance varied significantly among themes ($p < .001$).

\subsection{Reproducibility of Responses}
Table \ref{tab:tab1} demonstrated that DeepSeek-V3 achieved substantially higher accuracy with highly reproducible responses compared to less consistent ones (35 of 43 [81.4\%] vs 1 of 7 [14.3\%],  $p < .05$).  This highlights a statistically significant association between response reproducibility and accuracy in DeepSeek-V3 outputs. These findings underscore the potential of reproducibility as an indicator of response reliability in advanced LLMs, providing further evidence that reproducibility can serve as a marker of correctness in model-generated outputs.

\begin{table}[ht]
    \centering
    \caption{Answer statistics categorized by reproducibility level}\label{tab:tab1}
    \renewcommand{\arraystretch}{1.5}
        \resizebox{\textwidth}{!}{
                \setlength{\tabcolsep}{1mm}{
    \begin{tabular}{l|c|c|c}
    \hline
    \textbf{Reproducibility Type} & \textbf{Questions (N)} & \textbf{Correct (N) }&\textbf{Accuracy (\%, \boldmath
    $p<.05$)}\\ \hline \hline
    High Reproducibility  & 43 & 35 & 81.4 \\ \hline
    Low Reproducibility  & 7 & 1 & 14.3 \\ \hline
    \end{tabular}}}
\end{table}

\section{Discussion}  
\subsection{Findings} DeepSeek has garnered considerable attention due to its real-time response capabilities, logical consistency, and personalized dialogue interactions. Despite high expectations for LLMs in computer science education, comprehensive evaluations of their practical impact are lacking. Our study provides an empirical analysis of DeepSeek-V3, revealing the capability of DeepSeek-V3 in passing the CCNA test and the Network Engineer exam, along with the following core findings:  

\begin{description}
   
    \item[Finding 1] The Role Dependency experiments indicated that the model maintained consistent accuracy and reliability regardless of involving specific role definitions, highlighting its robustness and its capacity to capture contextual information without a predefined role \cite{liu2024performance}. These findings provide an empirical support for streamlining prompt engineering by simplifying role-specific information, which can reduce design costs, ease the interaction process, and enhance the overall response speed and sensitivity. Furthermore, this robustness offers insights into the optimization of context structure and semantic associations in prompts, facilitating more efficient prompt engineering. 
    
    \item[Finding 2] Cross-language examinations in Chinese and English exhibit marginally higher accuracy compared to Chinese only assessments, although the difference is statistically insignificant. This observation aligns with findings from GPT-4's performance on medical licensing tests \cite{liu2024performance,ting2024performance,fang2023does}. Notably, large language models demonstrate substantial performance variations across different domains and examination languages. \cite{alfertshofer2024sailing} analyzed ChatGPT's performance on medical licensing examinations across seven countries, revealing the highest accuracy (73\%) in Italian and the lowest (22\%) in French, with potential correlations to linguistic variations in question length. Furthermore, employing the methodology outlined in \cite{liu2024performance,fang2023does}, Chinese questions were translated into English for evaluation, yielding a slight improvement in accuracy—a pattern consistent with GPT-4's performance on translated medical examination questions \cite{liu2024performance,ting2024performance,fang2023does}.
    
    \item[Finding 3] The Question Type Performance experiment revealed that while the model demonstrated strong performance on lower-order factual recall tasks, it showed notable limitations in handling higher-order reasoning questions, suggesting room for improvement in its inferential capacities. Multiple-choice questions, which demand multi-step reasoning, proved significantly more challenging for the LLM compared to single-choice questions, resulting in substantially lower accuracy \cite{liu2024performance,mendoncca2024evaluating,bommasani2023holistic}. This disparity highlights the model's current constraints in processing complex question formats.
    
    \item[Finding 4] The Topic Performance experiment demonstrated that while the model achieved high accuracy in security fundamentals across both assessments, performance varied significantly across different topics in the Network Engineer exam, indicating a need for improved handling of complex domain-specific content. DeepSeek-V3 exhibited particular weakness in addressing network-related problems, including IP Connectivity, Network Access, and Network Fundamentals—a limitation consistent with observations in other LLMs \cite{donadel2024can}. This performance gap likely stems from the fact that such topics require not only theoretical knowledge but also practical expertise. Network construction and architecture evaluation present greater complexity and require more reasoning compared to purely theoretical questions.
    
    \item[Finding 5] The Response Reproducibility experiment demonstrated that when the model's responses are highly consistent, its accuracy is markedly improved. This result implies that response consistency not only affects output correctness but also reflects the model's inherent stability and quality. In other words, the reproducibility of LLM's answers can potentially serve as a metric for evaluating the reliability of advanced language models \cite{fang2023does}.  

\end{description}

\subsection{Limitations}
This study has the following limitations. Firstly, the number of test questions is relatively limited, and the constraint of sample size may weaken the wide applicability of the results. Secondly, the study lacked comparisons with human performance benchmarks, which could have provided more insight into the model's relative capabilities. Additionally, a more comprehensive assessment of the model's ability to handle multimodal data was not possible due to the exclusion of questions involving images from the study. 

\section{Conclusions}
In summary, this work presents a comprehensive assessment of DeepSeek-V3's capabilities and limitations in computer network security applications. Our evaluation of both the Network Engineer certification and CCNA examination reveals that while the model demonstrates strong performance in lower-order factual recall tasks, it exhibits measurable limitations when handling higher-order reasoning problems. This performance differential offers valuable insights for future model refinement. Furthermore, the model maintains consistent accuracy across Chinese and English question sets, confirming its robustness in cross-lingual scenarios. Our experiments show that variations in prompt design yield no statistically significant differences in model outputs, indicating inherent stability against input modifications. Notably, we introduce response repeatability as a novel metric for assessing output reliability, establishing its strong correlation with accuracy—a finding that suggests promising directions for future evaluation methodologies. This study extends the boundaries of AI applications in network security, while practically providing actionable methodologies for enhancing AI tool accuracy in professional certification contexts. These contributions demonstrate both immediate applicability and long-term research potential in the field.

%
%
%
 \bibliographystyle{splncs04}
 \bibliography{manuscript}
%







\end{document}